\documentclass[a4paper,conference]{IEEEtran}
\usepackage{cite}
\usepackage{multirow}
\usepackage{multicol}
\usepackage{amsmath,amssymb,amsfonts}
\usepackage{algorithmic}
\usepackage{graphicx}
\usepackage{textcomp}
\usepackage{xcolor}
\definecolor{RawSienna}{cmyk}{0,0.72,1,0.45}
\usepackage{graphicx}
\usepackage{comment}
\graphicspath{./images}
\def\BibTeX{{\rm B\kern-.05em{\sc i\kern-.025em b}\kern-.08em
    T\kern-.1667em\lower.7ex\hbox{E}\kern-.125emX}}
\begin{document}
%
\title{MixNet for Generalized Face Presentation Attack Detection}

\author{
\IEEEauthorblockN{Nilay Sanghvi$^{1}$, Sushant Kumar Singh$^{1}$, Akshay Agarwal$^{1,2}$, Mayank Vatsa$^{3}$, and Richa Singh$^{3}$}
\IEEEauthorblockA{$^{1}$IIIT-Delhi, India; $^{2}$Texas A\&M University, Kingsville, USA; $^{3}$IIT Jodhpur, India\\ $^{1}$\{nilay16063, sushant16103, akshaya\}@iiitd.ac.in; $^{3}$\{mvatsa, richa\}@iitj.ac.in
}
}


%


\maketitle

\begin{abstract}

The non-intrusive nature and high accuracy of face recognition algorithms have led to their successful deployment across multiple applications ranging from border access to mobile unlocking and digital payments. However, their vulnerability against sophisticated and cost-effective presentation attack mediums raises essential questions regarding its reliability. In the literature, several presentation attack detection algorithms are presented; however, they are still far behind from reality. The major problem with existing work is the generalizability against multiple attacks both in the seen and unseen setting. The algorithms which are useful for one kind of attack (such as print) perform unsatisfactorily for another type of attack (such as silicone masks). In this research, we have proposed a deep learning-based network termed as \textit{MixNet} to detect presentation attacks in cross-database and unseen attack settings. The proposed algorithm utilizes state-of-the-art convolutional neural network architectures and learns the feature mapping for each attack category. Experiments are performed using multiple challenging face presentation attack databases such as SMAD and Spoof In the Wild (SiW-M) databases. Extensive experiments and comparison with existing state of the art algorithms show the effectiveness of the proposed algorithm.

\end{abstract}


%
\IEEEpeerreviewmaketitle

\section{Introduction}
\label{sec:introduction}

Face being a non-intrusive biometrics modality has been deployed to various security-related areas ranging from constrained scenarios such as mobile unlocking to unconstrained scenarios such as surveillance. A forecast\footnote{https://www.mordorintelligence.com/industry-reports/facial-recognition-market} shows the popularity of face recognition, which claims that the face recognition market will increase to USD $10.9$ billion by $2025$ as compared to USD $4.4$ billion in $2019$. However, the significant challenge of the technology is the vulnerability against presentation attacks. For instance, an attacker can hide the identity by merely wearing a mask \cite{jia2020survey}, or an intruder can illegally access the system using a 2D printed photo \cite{bharadwaj2013computationally}. The tremendous amount of face images on social media platforms and unrestricted access to them can make it accessible to perform the attack. 

\begin{figure}
\centering
    \includegraphics[width=0.925\linewidth]{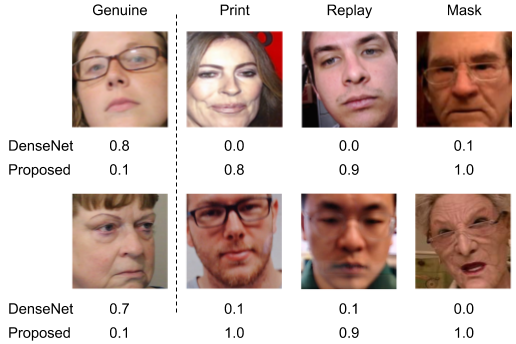}
    \caption{Images of genuine and different types of attack classes. The classification scores computed using DenseNet121 \cite{8099726} and the proposed algorithm are also written. For genuine class the score should be close to 0 and for attack it should be close to 1 for correct classification.}
    \label{fig:attacks}
\end{figure}

The prevalent presentation attacks on face recognition can be broadly classified into 2D artifacts based and 3D artifacts based attacks. 2D attacks cover printed photos using the printer and replay of photos or videos on an electronic screen. 3D attacks such as silicone masks and latex masks are sophisticated attacks and exhibit properties similar to the natural face. In the literature, several presentation attack detection (PAD) algorithms are presented, which are found effective in handling similar domain attacks, i.e., where the detector has seen the attack type or database at the time of training. However, the generalizability against multiple attacks is still a challenging task. At the same time, the development of new sophisticated silicone mask based attacks increases the detection complexity. The effectiveness of these 3D silicone masks can be seen in the following two cases: (i) a young person fooled the airport authority by wearing a mask and boarded the airplane\footnote{https://tinyurl.com/u4qda65} and (ii) face recognition algorithm in iPhone X is fooled by cost-effective masks\footnote{https://bgr.com/2017/11/28/face-id-hack-3d-mask-iphone-x-security/}.
On the other hand, 2D based attacking mediums can also be used for illegal access in unattended recognition systems. Therefore, the generalizability of the PAD algorithms across attack types is crucial. The aim of an effective PAD algorithm is to classify the images as genuine or attack in the first step so that the fake data is not processed through the recognition system. 


The prime objective of this research is to develop a generalized PAD algorithm. For an effective PAD algorithm, a challenging and unconstrained database is the first necessity. While several databases are presented in the literature, the significant limitation is the amount of data against each attack category. To address this problem, we have merged two challenging databases, namely SMAD \cite{7867821} and SiW-M \cite{liu2019deep}. The SMAD database contains silicone mask attack and authentic images captured in unconstrained settings. The SiW-M database is captured in the wild containing multiple attack types, including full masks and half masks. Further, we have also utilized two popular 2D attack databases, namely Replay-Attack \cite{chingovska2012effectiveness} and MSU-MFSD \cite{wen2015face} for experimentation and comparison with existing works. The experiments are performed in challenging conditions, including seen attacks, unseen attacks, and cross-database settings. As shown in Fig. \ref{fig:attacks}, the proposed algorithm is not only able to classify different images correctly, but also able to handle variations such as pose and illumination.

In brief, the key highlights of this research are:

\begin{itemize}
    \item A novel face PAD algorithm termed as MixNet is proposed. It consists of three sub-architectures, one each for detecting the three broad face presentation attacks - print, replay, and mask attack;
    \item The proposed algorithm, unlike existing algorithms, can further identify the type of attack images, i.e., whether the images come from print, replay or mask attack without an extra computational overhead;
    \item Extensive experiments concerning seen and unseen domains showcase the strength of the proposed algorithm as compared to hand-crafted features based and convolutional neural network (CNN) based PAD algorithms. 
\end{itemize}

\begin{figure*}[t]
    \centering
    \includegraphics[width=1.0\linewidth]{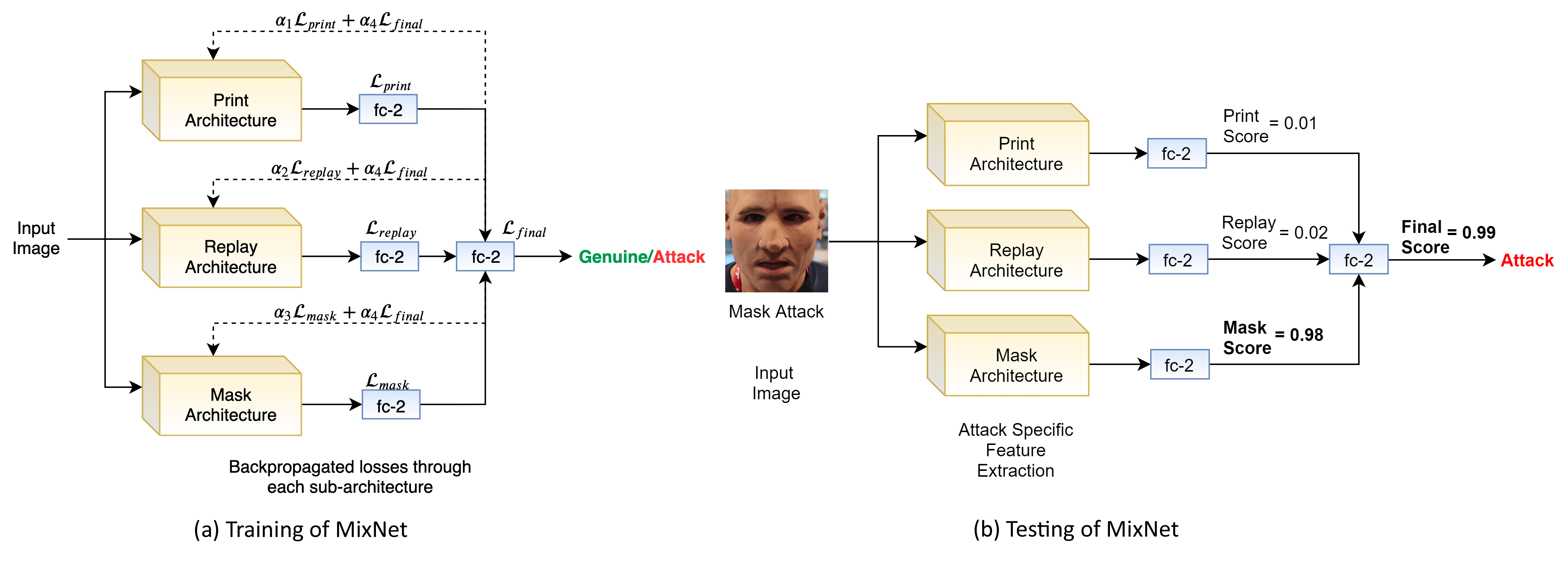}
    \caption{Schematic diagram of the proposed MixNet for face presentation attack detection.}
 \label{fig:mixnet}
\end{figure*}

\section{Related work}

The popular face PAD algorithms can be broadly grouped into pre-deep learning era and post-deep learning era. The pre-deep learning era based algorithms are mainly based on the extraction of texture features \cite{gragnaniello2015investigation,maatta2011face,agarwal2016face,agarwal2019btas,peng2018face,8014774}, motion cue based \cite{chingovska20132nd,komulainen2013complementary}, and hybrid algorithms \cite{siddiqui2016face,bharadwaj2013computationally,peng2020face}. The hand-crafted features based algorithms are computationally efficient and effective for the same domain attacks but lack generalizability against unseen attacks, databases, or even sensors. Moreover, generalizability is not the only issue, as shown by Agarwal et al. \cite{agarwal2019deceiving,agarwal2019deceiving2}, existing PAD algorithms can be fooled through feature manipulation or image transformations.

Menotti et al. \cite{menotti2015deep} and Tu and Fang \cite{tu2017ultra} proposed the deep architecture either through optimization or transfer learning to utilize them for face anti-spoofing. Liu et al. \cite{liu2019deep} have proposed deep tree learning for zero-shot attack detection. Recently, Mehta et al. \cite{mehta2019crafting} developed the panoptic face PAD algorithm using a shallow CNN model trained using focal loss. Their algorithm yields high detection accuracy on the individual attack and combined attack databases but lacks generalizability against unseen attack or database. Jia et al. \cite{jia2020face} have studied popular features implemented so far for PAD in mobile scenarios using multiple challenging databases. It is found that ResNet50 based detector yields the best result under cross-database testing. Furthermore, several survey papers have discussed existing PAD algorithms along with their limitations \cite{jia2020survey,singh2020robustness}.

In the literature, it is observed that most of the existing algorithms are useful for a particular kind of attack or database, but are less effective against multiple attacks in generalized settings. Therefore, in this research, with the aim of generalizability across multiple attacks, both in seen and unseen settings, a novel CNN architecture based PAD algorithm is presented. For each broad category of attack, a CNN architecture is deployed for feature learning and at the end confidence scores are combined together to yield the final detection result.

\section{Proposed Face PAD Algorithm: MixNet}

As explained in Section \ref{sec:introduction}, face presentation attacks can be broadly classified into three categories: print, replay, and 3D mask attacks. However, recent databases also contain the variations of these attacks such as half masks, paper masks, and transparent masks. An effective face PAD algorithm should be agnostic to these variations while detecting the traditional presentation attacks.



Most of the current algorithms have posed face PAD as a binary classification problem, and the algorithms learn to differentiate only between genuine and not genuine (i.e., attack) samples. It might be the reason because of which most of the existing algorithms are not generalized against unseen attacks. The characteristics of the print attack (hard surface, glossy, 2D) are entirely different from that of mask attack (smooth texture, 3D, similar to the skin); therefore, learning a single unified network is challenging. In the proposed algorithm, termed as MixNet, we have added an intermediate step of detecting the three broad attacks before the final classification of genuine/attack. MixNet consists of three sub-architectures where each of them learns the feature mapping of one of the three broad attacks.
\subsection{Training of MixNet}

On passing an attack sample to MixNet, only the sub-architecture responsible for detecting that attack should output a score close to 1. In contrast, the other two sub-architectures should output a score close to 0. Finally, after combining these three scores, MixNet shall return the final classification score close to 1 to denote the detection of an attack. To enforce the above process while training the architecture, we use four losses and label each data sample as a quadruple, which we explain in the following subsections \ref{sec:loss_function}, \ref{sec:architecture_details}.

\subsubsection{Loss Function}
\label{sec:loss_function}

Each sub-architecture has a loss associated with it, which results in three losses - \textit{`print loss'}, \textit{`replay loss'}, and \textit{`mask loss'}. A particular loss enforces the corresponding sub-architecture to detect the associated attack with high efficiency.
Further, there is \textbf{final classification loss} for the output layer (softmax) of MixNet to classify a sample as genuine or an attack. 
During training, MixNet tries to minimize the total loss: 
\begin{equation}
    \mathcal{L}_{total} =  \alpha_1\mathcal{L}_{print} + \alpha_2\mathcal{L}_{replay} + \alpha_3\mathcal{L}_{mask}
    + \alpha_4\mathcal{L}_{final}
    \label{eq:total_loss}
\end{equation}
where $\alpha_1, \alpha_2, \alpha_3, \alpha_4$ are the regularization coefficients for the four losses. Each of these losses is categorical cross-entropy loss represented as: \begin{equation}
    \mathcal{L}_{cross-entropy} = -\displaystyle\sum_{i} y_i \log(p_i) 
    \label{eq:loss_cce}
\end{equation}



Fig. \ref{fig:mixnet}(a) shows the forward and the backward pass of the proposed MixNet. When an image is forward passed, it goes through each architecture and based on the label of the input image, the amount of loss corresponding to each branch is back-propagated.   
Each loss affects only the layers that connect the input to the loss. For example, during backpropagation \textit{`print architecture'} would only be affected by \textit{`print loss'} and \textit{`final classification loss'}, i.e.,
\begin{equation}
    \alpha_1\mathcal{L}_{print}
    + \alpha_4\mathcal{L}_{final}
\end{equation}

\subsubsection{Architecture Details}
\label{sec:architecture_details}


For training the proposed MixNet, we label each data sample as described in Table \ref{table:data_labelling}. The first three entries correspond to the desired output from the three sub-architectures (print, replay, and mask architectures). On the other hand, the last entry corresponds to the final classification output of MixNet.

\begin{table}[b]
\centering
    \caption{Labeling of the input data for training MixNet.}
    \label{table:data_labelling}
   \setkeys{Gin}{keepaspectratio}
\resizebox*{0.49\textwidth}{0.49\textheight} {
    \begin{tabular}{|c|c|c|c|c|}
    \hline
    \textbf{Type of Sample} & \textbf{Print Label} & \textbf{Replay Label} & \textbf{Mask Label} & \textbf{Final Label}\\ 
    \hline
    Genuine & 0 & 0 & 0 & 0\\
    \hline
    Print Attack & 1 & 0 & 0 & 1\\
    \hline
    Replay Attack & 0 & 1 & 0 & 1\\
    \hline
    Mask Attack & 0 & 0 & 1 & 1\\
    \hline
    \end{tabular}
    }
\end{table}

The recent PAD algorithms \cite{jia2020face,wang2020unsupervised,Zhang2018CASIASURFAD} yields state-of-the-art performance when deep CNNs such as ResNet50 are used as the base network. Inspired from these studies, in this research, we have used three different deep CNN models for the three sub-architectures: ResNet50 \cite{he2015deep} (pre-trained on ImageNet \cite{5206848}), ResNet50-VF2 (pre-trained on VGGFace2 \cite{cao2017vggface2}) and DenseNet121 \cite{8099726} (pre-trained on ImageNet). The MixNet with these architectures are referred MixNet-ResNet50, MixNet-ResNet50-VF2 and MixNet-DenseNet121, respectively. The regularization coefficients for each network are as follows: (i) MixNet-ResNet50: $\alpha_1$ = 0.3, $\alpha_2$ = $0.5$,  $\alpha_3$ = $1.0$, $\alpha_4$ = $5.0$; (ii) MixNet-DenseNet121 and MixNet-ResNet50-VF2: $\alpha_1$ = $0.33$, $\alpha_2$ = $0.33$, $\alpha_3$ = $0.33$, $\alpha_4$ = $5.0$. The coefficients were experimentally found using grid-search ($\alpha_{1..3}=[0,1]$ and $\alpha_4 = [0,10]$) on the training set. The networks are trained with a batch size of $16$ to optimize the loss mentioned in section \ref{sec:loss_function} using SGD optimizer with the learning rate of $0.01$.

\subsection{Testing of MixNet}

Once the MixNet is trained, it is utilized for detecting different attacks and genuine samples. \textit{`print architecture'} would learn to detect print attacks. Similarly, \textit{`replay architecture'} and \textit{`mask architecture'} would learn to detect replay and mask attacks, respectively. 
Each sub-architecture outputs a score between 0 and 1, which indicates the confidence that the corresponding attack is present in the input image. For the final classification, MixNet combines the scores from the three sub-architectures. As shown in Fig. \ref{fig:mixnet}(b), when the input is mask attack image, mask architecture yields a score close to 1 while print and replay architectures output a score close to 0. In the end, the final softmax layer yields the score close to 1, which implies the input is an attack image. 


\begin{figure*}[t]
    \centering
    \includegraphics[width=0.85\linewidth]{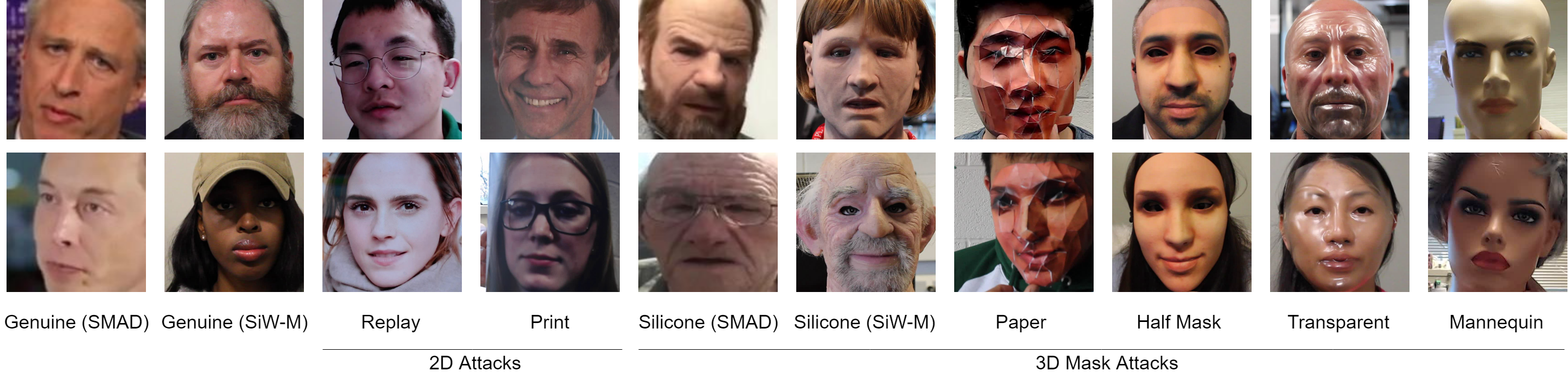}
    \caption{The examples of genuine and attack images from the merged database.}
    \label{fig:database_samples}
\end{figure*}

\section{Experimental Settings}
\label{sec:merged_dataset}

In this section, we describe the details of the experimental evaluations. Three popular presentation attacks, i.e., print, replay, and 3D mask, are utilized in our experiments. Print and replay attacks are cost-effective and simple to perform but are not as effective as 3D mask attacks, especially silicone masks, which are significantly costlier and developed using sophisticated hardware and software. For the efficacy of the proposed algorithm we have used two challenging databases; namely, Silicone Mask Attack Database (SMAD) and Spoof In the Wild with Multiple Attack Types (SiW-M). We have merged these databases together to create a large scale database effective for training. Next, each database is described followed by the description of experimental protocols and evaluation metrics used to report the results.  


\subsection{Databases}

\noindent\textbf{SMAD}: Manjani et al. \cite{7867821} created this first real-life silicone mask attack database consisting of a total of $130$ genuine and mask attack videos. Amongst the 65 real and 65 attack videos, $43$ and $59$ are males, respectively; rest belong to females. The authors have collected these videos from multiple sources on the web. Thus, it has many variations in background, illumination, facial expression, and video quality, making it a challenging database. 

\noindent\textbf{SiW-M}: Liu et al. \cite{liu2019deep} introduced SiW-M, which contains a total of $1,630$ videos of 5-7 seconds each. It consists of 968 videos of $13$ different attack types and $660$ authentic videos from $493$ subjects. The attack types include print attack, replay attack, and five types of mask attacks. The videos are captured with the variations in pose, lighting, and expression. SiW-M contains only $27$ videos of $12$ subjects for the silicone mask attack. To include sufficient silicone mask attack videos in our database while focusing on prevalent 2D and 3D attacks, we have merged print, replay, and five types of 3D mask attacks from the SiW-M database with SMAD. Fig. \ref{fig:database_samples} shows sample images from the above mentioned merged database.


\begin{table}[!t]
 \caption{Details of intra-database protocol. This is used to train and evaluate a model using three-fold cross-validation.}
\label{table:intra_database}
\centering
\begin{tabular}{|c|c|c|}
\hline
\textbf{Video Type} & \textbf{Database} & \textbf{Number of Videos}\\ 
\hline
Genuine & SMAD & 65\\
\hline
Genuine (from train split) & SiW-M & 217\\
\hline
Print Attack & SiW-M & 104\\
\hline
Replay Attack & SiW-M & 99\\
\hline
Mask Attack & SMAD & 65\\
\hline
\end{tabular}
\end{table}

\subsection{Experimental Protocols}
\label{sec:protocol}
We divide the merged database into two non-overlapping parts. For each part, we define a frame-based (classifying every single frame as attack or genuine) protocol.

\noindent \textbf{Intra-database Protocol}: This part contains the print and the replay attack videos from SiW-M and all the silicone attack videos from SMAD. For genuine data, we use all the $65$ genuine videos from SMAD and $217$ videos from the train split of the genuine data in SiW-M. We perform three-fold cross-validation in our experiments. Videos from each class (genuine, print, replay, and mask) of intra-database are equally divided into three non-overlapping folds. In each iteration of cross-validation, the model is trained on two folds and tested on the third fold. Table~\ref{table:intra_database} summarizes the details of this protocol.

\noindent{\textbf{Cross and Unseen Attack Protocol:}}
\label{sec:protocol}
This part is used to emulate cross-database testing and it helps evaluate the performance on unseen attacks. It includes all the five 3D mask attack videos of SiW-M and 131 genuine videos from the test split of the genuine data in SiW-M. Table~\ref{table:cross_database} presents the details of this protocol.
The three trained models, each from the three iterations of cross-validation performed in the intra-database protocol, are evaluated on this part. The results are reported as the average for the three models.
Testing on silicone mask attack videos of SiW-M emulates a cross-database scenario since the models are trained only on silicone mask videos from SMAD. Further, the other four 3D mask attacks: paper mask, half mask, transparent mask, and mannequin head are unseen attacks since the three trained models have not seen these attacks during training. We expect that since the training data has silicone mask attack samples, our proposed architecture should be able to generalize on these similar but unseen mask attacks.

\begin{table}[!t]
\caption{Details of cross and unseen attack database protocol. All the videos are from SiW-M. This is used to evaluate models trained using intra-database protocol under different scenarios. Seen: model has been trained on samples from the same database, Cross: model has been trained on similar samples from another database, Unseen: model has not been trained on these type of samples.}

\label{table:cross_database}
\centering
\begin{tabular}{|l|c|l|}
\hline
\textbf{Video Type} & \textbf{Number of Videos} & \textbf{Scenario}\\ 
\hline
Genuine (from test split) & 131 & Seen\\
\hline
Silicone Mask & 27 & Cross\\
\hline
Paper Mask & 17 & Unseen\\
\hline
Half Mask & 72 & Unseen\\
\hline
Transparent Mask & 88 & Unseen\\
\hline
Mannequin & 40 & Unseen\\
\hline
\end{tabular}
\end{table}





\subsection{Evaluation Metrics}
We have used the standard evaluation metrics defined by ISO/IEC 30107-3  \cite{ISO}: Receiver Operating Characteristic (ROC) curve, Attack Presentation Classification Error Rate (APCER) and Bona Fide Presentation Classification Error Rate (BPCER), and Average Classification Error Rate (ACER).
{\textbf{ROC}} is the plot of true positive rate (TPR) vs. false positive rate (FPR) calculated while varying the decision threshold for classification. The threshold for classification is computed on the Equal Error Rate (EER) of ROC, which is the error rate at the point where TPR equals the FPR. \textbf{APCER} is the fraction of presentation attack attempts that were successful and thus classified as genuine. \textbf{BPCER} is the fraction of bonafide samples falsely rejected as spoof. \textbf{ACER} is the average of APCER and BPCER. We report these metrics after averaging across the test sets of the three-fold cross-validation. In each iteration, we use the training fold to determine the threshold corresponding to the equal error rate (EER) and then use it to calculate ACER, APCER, and BPCER on the test sets.

\section{Experimental Results and Comparative Analysis}
\label{sec:experiments}
This section summarizes the experiments and their results to demonstrate the effectiveness of the proposed architecture. We use the Face Detector in Dlib library, which is a HOG+SVM based algorithm to crop face images from the videos of SiW-M and SMAD databases. We first describe the implementation details of the four algorithms used for comparison, then show their performance in the intra-database testing followed by the cross-domain testing results and analysis. For the cross-domain setting, the proposed algorithm is also compared with existing PAD algorithm, namely auxiliary supervision \cite{8578146}.


\subsection{Existing Algorithms}

The proposed algorithm is compared with two texture-based algorithms: LBP+HOG and Multi-scale LBP \cite{maatta2011face} and two deep learning algorithms: ResNet50 and DenseNet121. 


\subsubsection{Hand-Crafted Features + SVM}
In the first algorithm, we concatenate two popular hand-crafted features, namely HOG \cite{1467360} (Histogram of Oriented Gradients) and LBP  \cite{1017623} (Local Binary Patterns) histogram from an image, then apply SVM (Support Vector Machine) for classification. For LBP histogram, we obtain uniform non-rotation invariant $59$-bin histogram vector computed using $8$ sampling points on a circle of radius $1$. For HOG features, we use $9$ orientation bins, $16\times{16}$ pixels per cell, and apply L2-Hysteresis block normalization over blocks of $3\times3$ cells. Therefore, the dimension of the HOG features is $324$ (=$9\times{3}\times{3}\times{2}\times{2}$). In the second algorithm, we have used the formulation proposed in \cite{maatta2011face}, to calculate the multi-scale LBP histogram feature vector. The final histogram vector is passed to the non-linear SVM with radial basis function kernel for classification.


\subsubsection{Deep CNNs}
We have used ImageNet \cite{5206848} database trained CNNs, namely ResNet50 \cite{he2015deep} and DenseNet121 \cite{8099726} and fine-tuned for face PAD. The output layer of these CNNs are replaced by a fully connected softmax layer of 2 nodes representing real and attack class. 
The networks are fine-trained using stochastic gradient descent (SGD) to optimize categorical cross-entropy loss. The batch size and initial learning rate are set to $56$ and $0.01$, respectively.  
    
   
   
   

\begin{table}[!t]
    \centering
    \caption{Results (\%) in terms of $\mu$ $\pm$ $\sigma$ for intra-database protocol. Top-2 results are in bold.}
    \setkeys{Gin}{keepaspectratio}
\resizebox*{0.49\textwidth}{0.49\textheight} {
    \begin{tabular}{|l|c|c|c|} \hline
    \textbf{Architecture} & \textbf{ACER} & \textbf{APCER} & \textbf{BPCER}\\ 
    \hline 
    LBP+HOG & 14.98 $\pm$ 2.90 & 14.99 $\pm$ 6.15 & 14.96 $\pm$ 4.81\\ \hline
    
    Multi-scale LBP \cite{maatta2011face} & 16.01 $\pm$ 1.64 & 12.60 $\pm$ 0.65 & 19.43 $\pm$ 3.06 \\ \hline 
    
    ResNet50 & 10.05 $\pm$ 2.82 & 10.91 $\pm$ 5.21 & 9.18 $\pm$ 5.43\\ \hline
    
    MixNet-ResNet50 & 6.41 $\pm$ 0.69 & \textbf{2.34 $\pm$ 1.34} & 10.49 $\pm$ 2.06\\ \hline
    
    MixNet-ResNet50-VF2 & 6.85 $\pm$ 2.89 & 7.24 $\pm$ 4.46 & \textbf{6.47 $\pm$ 1.86}\\ \hline  
    
    DenseNet121 & \textbf{6.02 $\pm$ 0.63} & 7.16 $\pm$ 2.61 & \textbf{4.88 $\pm$ 3.87}\\ \hline
    
    MixNet-DenseNet121 & \textbf{4.52 $\pm$ 0.90} & \textbf{1.76 $\pm$ 1.31} & 7.28 $\pm$ 1.61 \\ \hline
    
    \end{tabular}
    }
    \label{table:results_intra}
\end{table}

\begin{figure}[!t]
\centering
    \includegraphics[width=0.8\linewidth]{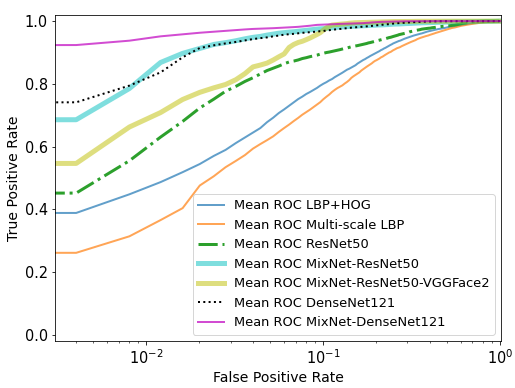}
    \caption{ROC for intra-database protocol.}
    \label{fig:roc}
\end{figure}

\subsection{Results for Intra-Database Protocol}

The intra-database protocol described in section \ref{sec:protocol} is followed for obtaining the results. For each algorithm, we have performed three-fold cross-validation. The error rates corresponding to intra-database protocol are reported in Table \ref{table:results_intra}.

The combination of LBP and HOG yields an average error rate of $14.98$\%, whereas, the LBP features computed over multiple scales yields the highest error rate among all the algorithms. The algorithms based on CNN models outperform the hand-crafted features based algorithms. Among the two CNN models used, the deeper model with 121 layers shows the lowest error rates for APCER and BPCER. However, the proposed MixNet-DenseNet121 reduces the attack error (i.e., APCER) and overall error (i.e., ACER) by $75.42$\% and $24.92$\%, respectively. On the other hand, the proposed MixNet with the ResNet50 model shows an improvement of $36.22$\% in the average detection error rate. It is interesting to note that DenseNet121 shows the lowest BPCER value. This could be because vanilla DenseNet and ResNet only have to classify samples as genuine or not genuine. In contrast, their corresponding MixNet versions focus on detecting the three types of attack. The MixNet corresponding to ResNet model pre-trained on face images shows a slightly higher error rate than object images based counterpart. Fig. \ref{fig:roc} shows the ROC curves for intra-database experiments obtained using different PAD algorithms.

\textbf{Attack-wise APCER:} We have also performed the ablation study to see whether simultaneous learning in MixNet helps in detecting a specific attack. The 3 fold cross-validation experimental results on merged database shows that the proposed MixNet yields APCER value of $0.0 \pm {0.0}$\%, $0.08 \pm {0.12}$\%, and $2.87 \pm 2.06$\% for print, replay, and mask attack respectively. On the other hand, separately training three DenseNet121 each dedicated to classify a specific attack and taking the maximum of their output scores gives APCER of $2.94 \pm {2.03}$\%, $5.74 \pm {4.22}$\%, and $6.71 \pm {0.50}$\% for print, replay, and mask attack, respectively. If we take the average of output scores, the APCERs are $1.97 \pm {2.36}$\%, $3.70 \pm {4.25}$\%, and $7.65 \pm {3.80}$\% for print, replay, and mask attack, respectively. Finally, vanilla DenseNet121 yields APCER of $1.59 \pm {0.85}$\%, $1.68 \pm {1.45}$\%, and $11.08 \pm {3.46}$\% for the three attacks, respectively. The lower error rate for each attack showcases the advantage of proposed MixNet over simpler learning. 





\begin{table}[!t]
    \caption{Results (\%) in terms of $\mu$ $\pm$ $\sigma$ on cross and unseen attack protocol. Top-2 results are in bold.}
    \label{table:overall_results_cross}
    \centering
    \setkeys{Gin}{keepaspectratio}
\resizebox*{0.49\textwidth}{0.49\textheight} {
    \begin{tabular}{|l|c|c|c|}
    \hline
    \textbf{Architecture} & \textbf{ACER} & \textbf{APCER} & \textbf{BPCER}
    \\ \hline
    LBP+HOG & 35.68 $\pm$ 0.99 & 57.70 $\pm$ 0.65 & 13.66 $\pm$ 1.47 \\  \hline
    
    Multi-scale LBP \cite{maatta2011face} & 32.86 $\pm$ 0.90 & 51.55 $\pm$ 2.60 & 14.18 $\pm$ 2.03 \\ \hline
    
    ResNet50 & 35.48 $\pm$ 0.54 & 56.75 $\pm$ 2.32 & 14.21 $\pm$ 1.41\\ \hline
    
    MixNet-ResNet50 & \textbf{23.69 $\pm$ 4.77} & \textbf{35.48 $\pm$ 10.10} & \textbf{11.89 $\pm$ 0.96}\\
    \hline
    
    MixNet-ResNet50-VF2 & 32.95 $\pm$ 1.29 & 53.81 $\pm$ 3.27 & 
    12.10 $\pm$ 0.71\\
    \hline
    
    DenseNet121 & 30.84 $\pm$ 1.97 & 50.48 $\pm$ 4.24 & \textbf{11.20 $\pm$ 0.51}\\ \hline
    
    MixNet-DenseNet121 & \textbf{24.72  $\pm$ 0.61} & \textbf{36.93 $\pm$ 0.95} & 12.51 $\pm$ 0.64\\ \hline
    \end{tabular}
    }
\end{table}

\begin{table*}[!t]
\caption{APCER (\%) attack-wise for cross and unseen attack protocol. Top-2 results are in bold.}
\label{table:results_cross}
\centering
\begin{tabular}{|l|c|c|c|c|c|}
\hline
\textbf{Architecture} & \textbf{Silicone Mask} & \textbf{Paper Mask} & \textbf{Half Mask} &  \textbf{Transparent Mask} &  \textbf{Mannequin} \\ \hline
LBP+HOG & 53.33 & 21.44 & 54.06 & 83.85 & 26.98 \\ \hline

Multi-scale LBP \cite{maatta2011face} & 45.68 & 4.84 & 42.91 & 84.74 & 21.03 \\ \hline

ResNet50 & \textbf{12.84} & 97.50 & 44.28 & 98.18 & 31.32  \\ \hline

MixNet-ResNet50 & 16.22 &  \textbf{1.00} &  \textbf{26.41} &  \textbf{71.54} & 
\textbf{4.78} \\ \hline

MixNet-ResNet50-VF2 & 17.74 &  10.20 &  61.73 &  82.46 & 
23.67 \\ \hline

DenseNet121 & 23.12 & 26.12 & 39.92 & 92.94 & 9.34 \\ \hline

MixNet-DenseNet121 & \textbf{11.54} & \textbf{4.54} & \textbf{21.56} & \textbf{81.56} & 
\textbf{3.54} \\ \hline
\end{tabular}
\end{table*}

\subsection{Results for Cross and Unseen Attack Protocol}


As described in section \ref{sec:protocol}, the models corresponding to each fold trained on intra-database protocols are used to evaluate on cross-database and unseen attack settings. Table \ref{table:overall_results_cross} shows the average ACER, APCER, and BPCER for this protocol.
It is found that the proposed MixNet utilizing ResNet50 performs slightly better than the MixNet utilizing the DenseNet121. The PAD algorithms based on hand-crafted features and deep CNN models yield  APCER of at-least $50.48$\% while their BPCER is significantly lower. 
The proposed MixNet-ResNet50 reduces the error rates of ResNet50 by at-least $16.33$\%. Similarly, the MixNet-DenseNet121 reduces the error rates of the fine-tuned DenseNet121 model significantly. For example, the attack samples detection error rate of MixNet is $26.84$\% lower than the DenseNet121 model. Interestingly, we have observed that the MixNet with ResNet50 model pre-trained on VGGFace2 database yields higher ACER value as compared to the MixNet with ResNet50 model pre-trained on the ImageNet database. 

\par
Table \ref{table:results_cross} shows the attack detection error for each of the five 3D mask attacks. 
The hand-crafted algorithms which lack generalizability fail significantly for silicone mask and transparent mask. The paper mask is found to be the easiest attack to be detected, which might be because it lacks the smooth texture and suffers from edge artifacts. Even in such a case, the ResNet50 model shows more than $97$\% error rate. On unseen attack types such as silicone mask, paper mask, half mask, and mannequin, the error rate of the proposed MixNet-DenseNet121 is $11.54$\%, $4.54$\%, and $21.56$\%, and $3.54$\%, respectively. The effectiveness of the proposed algorithm on attacks such as mannequin, which is not explored previously in the literature, shows that it is generalizable to handle the real-world scenarios. On the easiest paper-based mask, the fine-tuned DenseNet121 model shows a $26.12$\% error rate, which is $82.62$\% higher than the MixNet. On the remaining attacks, the error rate of the proposed algorithm is at-least $12.24$\% lower than DenseNet121. Further, the performance of the proposed algorithm for mannequin attack detection is $78.4$\% and $7.9$\% better than \textbf{auxiliary supervision} \cite{8578146} and \textbf{deep tree} \cite{liu2019deep}, respectively. The silicone and half-mask detection error rates of the proposed MixNet-DenseNet121 are $36.8$\% and $22.0$\% lower than the auxiliary supervision \cite{8578146}, respectively. We have also observed that every model performed poorly on samples from a transparent mask. It may be because the real face behind a transparent mask is almost visible, closely resembling a natural face. However, the error rate of the proposed algorithm is $27.9$\% lower than auxiliary supervision approach \cite{8578146}.

\subsection{Visualization and Analysis}
\begin{figure}[t]
\centering
    \includegraphics[width=0.95\linewidth]{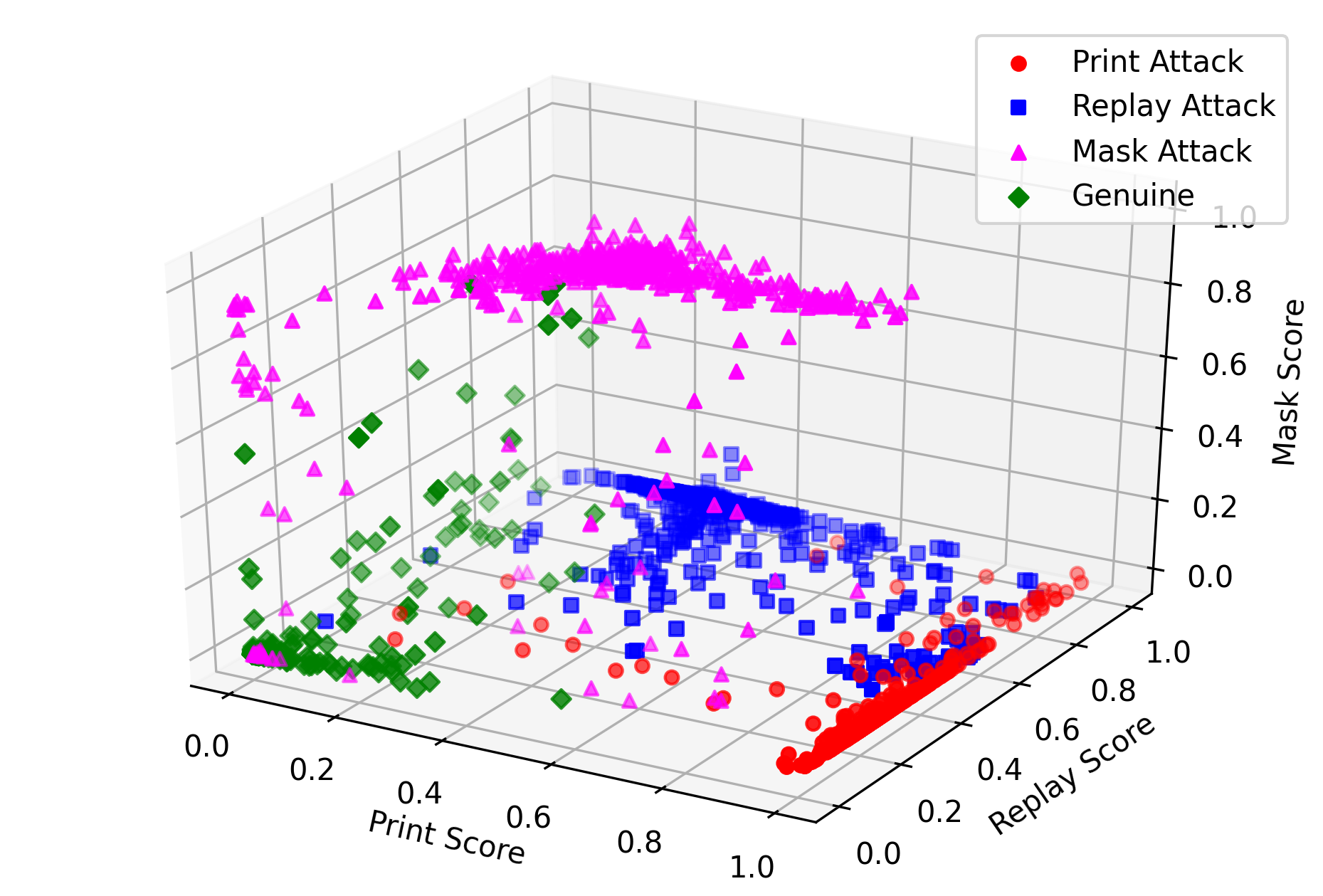}   
    \caption{Visualisation of output scores from the three sub-architectures of MixNet-DenseNet121 for a subset of test samples.}
    \label{fig:scatter_plot}
\end{figure}

A 3D scatter plot shown in Fig. \ref{fig:scatter_plot} is used to visualize the output scores from three sub-architectures of MixNet and to showcase their importance in detecting the particular attacks. It is observed that the samples of genuine class and the three attack classes form four separate clusters with minimal overlap. This shows that MixNet can not only detect an attack but even classify the type of the attack. It is observed that the scores of mask attack samples are intermixed with the cluster of genuine samples. In contrast, print and replay attack samples are distant from the cluster of genuine samples. It implies that mask attacks resemble real-life face texture and quality more than print and replay attacks. Thus, the major challenge lies in detecting mask attacks.


\begin{figure}[t]
\centering
    \includegraphics[width=0.825\linewidth]{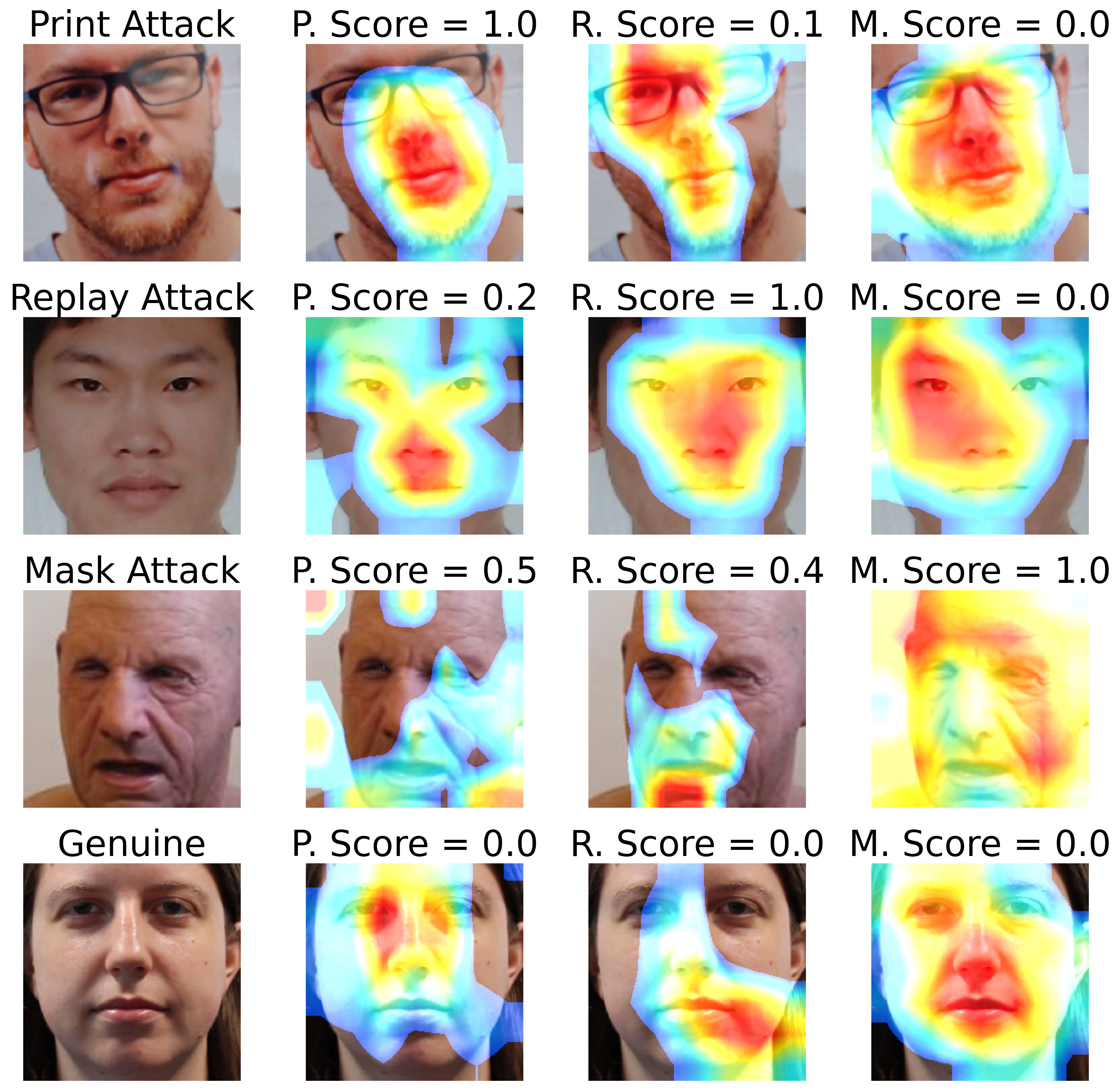}
    \caption{The Class Activation Maps (CAMs) of four kinds of images obtained from MixNet-DenseNet121. From left to right: original image, CAM of print, replay, and mask architecture, respectively. The CAMs highlight the image regions used by each sub-architecture to detect the corresponding attack. P. Score, R. Score, M. Score represent the scores of Print, Replay, and Mask class, respectively.}
    \label{fig:activation_maps}
\end{figure}

Deep Learning based techniques are often considered as black boxes. To gain an insight into what kind of features are learned by MixNet, we visualize the class activation maps (CAM) \cite{zhou2015learning}. CAM is used to highlight the regions in an input image that are most effective for classification. Fig. \ref{fig:activation_maps} provides some interesting insights: print and replay attacks are detected using the regions around the nose and mouth, while for the mask attacks, the full face region is required. The depth around the nose region is different from the rest of the face, and hence, print and replay attacks can be detected using these regions. Mask attacks, however, can be detected using the areas where there is an opening in the eyes and mouth and also by finding discriminative patterns on cheek and forehead regions.

\section{Ablation Study}

In this section, we showcase the performance of the proposed algorithm on existing face PAD databases using their predefined protocol. Other than that, in place of training the three sub-architectures simultaneously as MixNet, the performance of sub-architectures trained independently is also studied.


\textbf{Results on Existing Databases:} We have performed experiments on the existing databases, namely Replay-Attack \cite{chingovska2012effectiveness} and MSU-MFSD \cite{wen2015face} for an extensive comparison of MixNet with other face PAD algorithms. \textbf{Replay-Attack } consists of 1200 real, print, and replay videos of $50$ subjects. \textbf{MSU-MFSD} has $280$ videos of photo and video attack attempts of $35$ subjects. We have used the predefined train test split of both the databases to make a fair comparison with the existing algorithm. Since Replay-Attack and MSU-MFSD do not contain mask attack samples, the MixNet-DenseNet for these experiments only had two sub-architectures, one each for print and replay attack. Table \ref{table:replay-attack} shows the face PAD results of the proposed and existing algorithms on Replay-Attack and MSU-MFSD. The results show the effectiveness of the proposed algorithm by surpassing several existing algorithms based on the fusion of classifiers, image regions, and features. For example, the recently proposed algorithm DR-UDA \cite{wang2020unsupervised} consisting of three modules: a source domain metric learning network (ML-Net), an unsupervised adversarial domain adaptation module (UDA-Net), and a disentangled representation learning module (DR-Net) achieves $1.3$\% HTER and $6.3$\% EER on the Replay-Attack and MSU database, respectively. On the other hand, the proposed MixNet improves the performance to $0.6$\% HTER and $0.4$\% EER on Replay-Attack and MSU-MFSD datasets, respectively.

\textbf{Simultaneous vs. Independent Sub-architectures Training:} We have also compared the proposed MixNet to a method where we separately train models, each dedicated to classifying a specific attack type. The final output score is optimized on the training or validation set using the maximum or average rule of these models' output scores. We selected the best model among Xception \cite{chollet2017xception}, DenseNet121 \cite{8099726}, and ResNet50 \cite{he2015deep} for each attack type based on HTER on the validation set. For Replay-Attack, the best settings were ResNet50 for the print attack model, DenseNet121 for the replay attack model, and the average score for final output. Similarly, for MSU-MFSD, the best configuration used DenseNet121 for both the attack types and took the maximum of these scores for the final output score. As shown in Table VII, the lowest test EER on MSU-MFSD is \textbf{2.36\%} which is $1.96$\% higher than the MixNet. Similarly, the lowest test HTER of the independent sub-architectures on Replay-Attack is \textbf{0.68\%} higher than the proposed MixNet.






\begin{table}[!t]
    \centering
    \caption{Intra-database evaluation on Replay-Attack (HTER\%) and MSU-MFSD (EER\%).}
    \setkeys{Gin}{keepaspectratio}
\resizebox*{0.49\textwidth}{0.49\textheight} {
    \begin{tabular}{|l|c|c|c|} 
    \hline
    \textbf{Method}&
    \textbf{Replay-Attack} & \textbf{MSU-MFSD }\\ 
    \hline

    Haralick Features \cite{agarwal2016face}&
    - &
    5.0 \\
    \hline
    

    
    
    
    
    Deep Learning \cite{wang2018unsupervised} &
    2.1 &
    5.8 \\
    
    \hline
    ResNet18 \cite{he2015deep} &
    2.8 &
    8.7 \\
    \hline
    
    DR-UDA (ResNet18) \cite{wang2020unsupervised} &
    1.4 &
    6.0 \\
    \hline
    
    SE-ResNet18 \cite{hu2018squeeze} &
    2.4 &
    8.7 \\
    \hline
    
    DR-UDA (SE-ResNet18) \cite{wang2020unsupervised} &
    1.3 &
    6.3 \\
    \hline
    Multi-Regional CNN \cite{MA2020} & 1.6 & - \\ \hline
    CCoLBP+Ensemble Learning  \cite{peng2020face} & 4.0 & 5.0 \\ \hline
    SfSNet \cite{pinto2020leveraging} & 3.1 & - \\ \hline
    Independently Optimized Sub-Nets & 1.3 & 2.4 \\ \hline
    \textbf{Ours (MixNet-DenseNet)} &
    \textbf{0.6} &
    \textbf{0.4}\\
    \hline
    
    \end{tabular}
    }
    \label{table:replay-attack}
\end{table}

\section{Conclusion and Future Work}

The vulnerability of facial recognition algorithms to presentation attacks limit their usability for security purposes. Thus, it becomes essential to develop more reliable and robust algorithms to detect such attacks on facial recognition. This paper introduces a novel architecture termed as \textbf{MixNet}, which utilizes three sub-architectures to identify the particular presentation attack. Experimental results show that MixNet outperforms multiple face PAD algorithms based on CNN architectures and hand-crafted features to detect seen and unseen attacks. 
Currently, for each sub-architecture in MixNet, the same network has been used. We plan to explore the selection of different architectures such that they are state of the art for detecting the corresponding attack. Finally, we believe the application of MixNet is not limited to face presentation attack detection but can also be extended to other biometrics such as iris and fingerprint PAD.

\section*{Acknowledgment}

A. Agarwal is partly supported by the Visvesvaraya PhD Fellowship. R. Singh and M. Vatsa are partially supported through a research grant from MeitY, India. M. Vatsa is also partially supported through the Swarnajayanti Fellowship by the  Government of India.






\bibliographystyle{IEEEtran}
\bibliography{1502}
\end{document}